\pdfoutput=1

\documentclass[11pt]{article}

\usepackage[]{EMNLP2022}

\usepackage{times}
\usepackage{latexsym}

\usepackage[T1]{fontenc}

\usepackage[utf8]{inputenc}

\usepackage{microtype}

\usepackage{inconsolata}

\usepackage{amsmath}
\usepackage{amssymb}
\usepackage{bbm}
\DeclareMathOperator*{\argmin}{arg\,min}
\usepackage{graphicx}
\usepackage{multirow}

\title{OTSeq2Set: An Optimal Transport Enhanced Sequence-to-Set Model for Extreme Multi-label Text Classification}

\author{Jie Cao \\
  Polytechnic Institute\\
  Zhejiang University\\
  \texttt{caojie@zju.edu.cn} \\\And
  Yin Zhang\Thanks{Corresponding author} \\
  College of Computer Science and Technology\\
  Zhejiang University\\
  \texttt{yinzh@zju.edu.cn} \\}

\begin{document}
\maketitle
\begin{abstract}
Extreme multi-label text classification (XMTC) is the task of finding the most relevant subset labels from an extremely large-scale label collection. Recently, some deep learning models have achieved state-of-the-art results in XMTC tasks. These models commonly predict scores for all labels by a fully connected layer as the last layer of the model. However, such models can't predict a relatively complete and variable-length label subset for each document, because they select positive labels relevant to the document by a fixed threshold or take top k labels in descending order of scores. A less popular type of deep learning models called sequence-to-sequence (Seq2Seq) focus on predicting variable-length positive labels in sequence style. However, the labels in XMTC tasks are essentially an unordered set rather than an ordered sequence, the default order of labels restrains Seq2Seq models in training. To address this limitation in Seq2Seq, we propose an autoregressive sequence-to-set model for XMTC tasks named OTSeq2Set. Our model generates predictions in student-forcing scheme and is trained by a loss function based on bipartite matching which enables permutation-invariance. Meanwhile, we use the optimal transport distance as a measurement to force the model to focus on the closest labels in semantic label space. Experiments show that OTSeq2Set outperforms other competitive baselines on 4 benchmark datasets. Especially, on the Wikipedia dataset with 31k labels, it outperforms the state-of-the-art Seq2Seq method by 16.34\% in micro-F1 score. The code is available at https://github.com/caojie54/OTSeq2Set.
\end{abstract}

\section{Introduction}
Extreme multi-label text classification (XMTC) is a Natural Language Processing (NLP) task of finding the most relevant subset labels from an extremely large-scale label set. It has a lot of usage scenarios, such as item categorization in e-commerce and tagging Wikipedia articles. XMTC become more important with the fast growth of big data.

As in many other NLP tasks, deep learning based models also achieve the state-of-the-art performance in XTMC. For example, AttentionXML \citep{you_attentionxml_2019}, X-Transformer \citep{chang_Taming_2020} and LightXML \citep{jiang_lightxml_2021} have achieved remarkable improvements in evaluating metrics relative to the current state-of-the-art methods. These models are both composed of three parts \citep{jiang_lightxml_2021}: text representing, label recalling, and label ranking. The first part converts the raw text to text representation vectors, then the label recalling part gives scores for all cluster or tree nodes including portion labels, and finally, the label ranking part predicts scores for all labels in descending order. Notice that, the label recalling and label ranking part both use fully connected layers. Although the fully connected layer based models have excellent performance, there exists a drawback which is these models can't generate a variable-length and relatively complete label set for each document. Because the fully connected layer based models select positive labels relevant to the document by a fixed threshold or take top k labels in descending order of label scores, which depends on human's decision. Another type of deep learning based models is Seq2seq learning based methods which focus on predicting variable-length positive labels only, such as MLC2Seq \citep{nam_maximizing_2017}, SGM \citep{yang_sgm_2018}. MLC2Seq and SGM enhance Seq2Seq model for Multi-label classification (MLC) tasks by changing label permutations according to the frequency of labels. However, a pre-defined label order can't solve the problem of Seq2Seq based models which is the labels in XMTC tasks are essentially an unordered set rather than an ordered sequence. \citet{yang_deep_2019} solves this problem on MLC tasks via reinforcement learning by designing a reward function to reduce the dependence of the model on the label order, but it needs to pretrain the model via Maximum Likelihood Estimate (MLE) method. The two-stage training is not efficient for XMTC tasks that have large-scale labels.

To address the above problems, we propose an autoregressive sequence-to-set model, OTSeq2Set, which generates a subset of labels for each document and ignores the order of ground truth in training. OTSeq2Set is based on the Seq2Seq \citep{bahdanau_neural_2016}, which consists of an encoder and a decoder with the attention mechanism. The bipartite matching method has been successfully applied in Named entity recognition task \citep{tan_sequence--set_2021} and keyphrase generation task \citep{ye_one2set_2021} to allievate the impact of order in targets. \citet{chen_improving_2019} and \citet{li_improving_2020} have successfully applied the optimal transport algorithm to enable sequence-level training for Seq2Seq learning. Both methods can achieve optimal matching between two sequences, but the difference is the former matches two sequences one to one, and the latter gives a matrix containing regularized scores of all connections. We combine the two methods in our model.

Our contributions of this paper are summarized as follows:
(1) We propose two schemes to use the bipartite matching in XMTC tasks, which are suitable for datasets with different label distributions. (2) We combine the bipartite matching and the optimal transport distance to compute the overall training loss, with the student-forcing scheme when generating predictions in the training stage. Our model can avoid the exposure bias; besides, the optimal transport distance as a measurement forces the model to focus on the closest labels in semantic label space. (3) We add a lightweight convolution module into the Seq2Seq models, which achieves a stable improvement and requires only a few parameters. (4) Experimental results show that our model achieves significant improvements on four benchmark datasets. For example, on the Wikipedia dataset with 31k labels, it outperforms the state-of-the-art method by 16.34\% in micro-F1 score, and on Amazon-670K, it outperforms the state-of-the-art model by 14.86\% in micro-F1 score.

\section{Methodology}
\subsection{Overview}
Here we define necessary notations and describe the Sequence-to-Set XMTC task. Given a text sequence $\boldsymbol{x}$ containing $l$ words, the task aims to assign a subset $\boldsymbol{y}^g$ containing $n$ labels in the total label set $\boldsymbol{L}$ to $\boldsymbol{x}$. Unlike fully connected layer based methods which give scores to all labels, the Seq2Set XMTC task is modeled as finding an optimal positive label sequence $\boldsymbol{y}^g$ that maximizes the joint probability $P(\boldsymbol{\hat{y}}|\boldsymbol{x})$, which is as follows:
\begin{equation}\label{eq:probability}
    P(\boldsymbol{\hat{y}}|\boldsymbol{x}) = \prod_{i=1}^{n}P(y_{\hat{\rho}{(i)}}|y^g_1,y^g_2,\ldots,y^g_{i-1},\boldsymbol{x}),
\end{equation}
where $\boldsymbol{y}^g$ is the sequence generated by the greedy search, $\boldsymbol{y}$ is the ground truth sequence with default order, $\boldsymbol{\hat{y}}$ is the most matched reordered sequence computed by bipartite matching. As described in Eq.(\ref{eq:probability}), we use the student-forcing scheme to avoid \textit{exposure bias} \citep{ranzato_sequence_2016} between the generation stage and the training stage. Furthermore, combining the scheme with bipartite matching enables the model to eliminate the influence of the default order of labels.



\subsection{Sequence-to-Set Model}
Our proposed Seq2Set model is based on the Seq2Seq \citep{bahdanau_neural_2016} model, and the model consists of an encoder and a set decoder with the attention mechanism and an extra lightweight convolution layer \citep{wu_pay_2019}, which are introduced in detail below.
\paragraph{Encoder}
We implement the encoder by a bidirectional GRU to read the text sequence $\boldsymbol{x}$ from both directions and compute the hidden states for each word as follows:
\begin{equation}
    \overrightarrow{\boldsymbol{h}}_i = \overrightarrow{\mathrm{GRU}}(\overrightarrow{\boldsymbol{h}}_{i-1}, \boldsymbol{e}(x_i))
\end{equation}
\begin{equation}
    \overleftarrow{\boldsymbol{h}}_i = \overleftarrow{\mathrm{GRU}}(\overleftarrow{\boldsymbol{h}}_{i+1}, \boldsymbol{e}(x_i))
\end{equation}
where $\boldsymbol{e}(x_i)$ is the embedding of $x_i$. The final representation of the $i$-th word is $\boldsymbol{h}_i = [\overrightarrow{\boldsymbol{h}}_i;\overleftarrow{\boldsymbol{h}}_i]$ which is the concatenation of hidden states from both directions.
\paragraph{Attention with lightweight convolution}
After the encoder computes $\boldsymbol{h}_i$ for all elements in $\boldsymbol{x}$, we compute a context vector $\boldsymbol{c}_t$ to focus on different portions of the text sequence when the decoder generates the hidden state $\boldsymbol{s}_t$ at time step $t$,
\begin{equation}
    \boldsymbol{c}_t = \sum_{i=1}^l \alpha_{ti}\boldsymbol{h}_i
\end{equation}
The attention score $\alpha_{ti}$ of each representation $h_i$ is computed by
\begin{equation}
    \alpha_{ti} = \frac{\exp{(e_{ti})}}{\sum_{j=1}^l\exp{(e_{tj})}}
\end{equation}
\begin{equation}
    e_{ti} = \boldsymbol{v}_a^{\top}\tanh{(\boldsymbol{W}_a\boldsymbol{s}_t+\boldsymbol{U}_a\boldsymbol{h}_i)}
\end{equation}
where $\boldsymbol{W}_a$, $\boldsymbol{U}_a$, $\boldsymbol{v}_a$ are weight parameters. For simplicity, all bias terms are omitted in this paper.

To maximally utilize hidden vectors $\{\boldsymbol{h}_i\}_{i=1,\ldots,l}$ in the encoder, we use the lightweight convolution layer to compute "label" level hidden vectors $\{\boldsymbol{\hat{h}}_i\}_{i=1,\ldots,k}$, then compute another context vector $\boldsymbol{\hat{c}}_t$ which uses the same parameters as $\boldsymbol{c}_t$,
\begin{equation}
    \{\boldsymbol{\hat{h}}_i\}_{i=1,\ldots,k} = \mathrm{LightConv}(\{\boldsymbol{h}_i\}_{i=1,\ldots,l}).
\end{equation}
The \textit{lightweight convolutions} are depth-wise separable \citep{wu_pay_2019}, softmax-normalized and share weights over the channel dimension. Readers can refer to \citet{wu_pay_2019} for more details.
\paragraph{Decoder}
The hidden state $\boldsymbol{s}_t$ of decoder at time-step $t$ is computed as follows:
\begin{equation}
    \boldsymbol{s}_t = \mathrm{GRU}(\boldsymbol{s}_{t-1},[\boldsymbol{e}(\boldsymbol{p}_{t-1});\boldsymbol{c}_t;\boldsymbol{\hat{c}}_t])
\end{equation}
where $\boldsymbol{e}(\boldsymbol{p}_{t-1})$ is the embedding of the label which has the highest probability under the distribution $\boldsymbol{p}_{t-1}$. $\boldsymbol{p}_{t-1}$ is the probability distribution over the total label set $\boldsymbol{L}$ at time-step $t-1$ and is computed by a fully connected layer:
\begin{multline}
    \boldsymbol{p}_t = \mathrm{softmax}(\boldsymbol{W}_p[\boldsymbol{s}_t;\boldsymbol{c}_t;\boldsymbol{\boldsymbol{\hat{c}}_t};\boldsymbol{e}(\boldsymbol{p}_{t-1})])
\end{multline}
where $\boldsymbol{W}_p\in \mathbbm{R}^{V\times z}$ is weight parameters.

The overall label size $V$ of the total label set $\boldsymbol{L}$ is usually huge in XMTC. In order to let the model fit into limited GPU memory, we use the hidden bottleneck layer like XML-CNN \citep{liu_deep_2017} to replace the fully connected layer. The bottleneck layer is described as follows:
\begin{multline}
    \boldsymbol{p}_t = \mathrm{softmax}(\boldsymbol{W}_{out}\\
    \tanh{(\boldsymbol{W}_{in}[\boldsymbol{s}_t;\boldsymbol{c}_t;\boldsymbol{\boldsymbol{\hat{c}}_t};\boldsymbol{e}(\boldsymbol{p}_{t-1})])})
\end{multline}
where $\boldsymbol{W}_{in}\in \mathbbm{R}^{b\times z}$, $\boldsymbol{W}_{out}\in \mathbbm{R}^{V\times b}$ are weight parameters, and the hyper-parameter $b$ is the bottleneck size. The size of the parameters in this part reduces from a vast size $O(V\times z)$ to a much smaller $O((V+z)\times b)$. According to the size of labels for different datasets, we can set different $b$ to make use of GPU memory.
\subsection{Loss Function}
\subsubsection{Bipartite Matching}
After generating N predictions in student-forcing scheme, we need to find the most matched reordered ground truth sequence $\boldsymbol{\hat{y}}$ by bipartite matching between ground truth sequence $\boldsymbol{y}$ with default order and the sequence  $\boldsymbol{p}$ of generated distribution. To find the optimal matching we search for a permutation $\hat{\rho}$ with the lowest cost:
\begin{equation}
    \hat{\rho} = \argmin_{\rho\in\mathcal{O}_N}\sum_{i=1}^{N}\mathcal{C}_{\mathsf{match}}(y_{\rho(i)},\boldsymbol{p}_i),
\end{equation}
where $\mathcal{O}_N$ is the space of all permutations with length $N$, $\mathcal{C}_{\mathsf{match}}(y_{\rho(i)},\boldsymbol{p}_i)$ is a pair matching cost between the ground truth label with index $\rho(i)$ and generated distribution $\boldsymbol{p}_i$ at time step $i$. We use the Hungarian method \citep{kuhn1955hungarian} to solve this optimal assignment problem. The matching cost considers the generated distributions and is defined as follows:
\begin{equation}
    \mathcal{C}_{\mathsf{match}}(y_{\rho(i)},\boldsymbol{p}_i) = -\mathbbm{1}_{\{y_{\rho(i)} \neq \varnothing\}}\boldsymbol{p}_{i}(y_{\rho(i)}),
\end{equation}
where $\boldsymbol{p}_{i}(y_{\rho(i)})$ is the probability of label $y_{\rho(i)}$. $y_{\rho(i)}\neq\varnothing$ means that the distributions only match with non-$\varnothing$ ground truth labels. 

\begin{figure}[t]
    \centering
    \includegraphics[width=7.5cm]{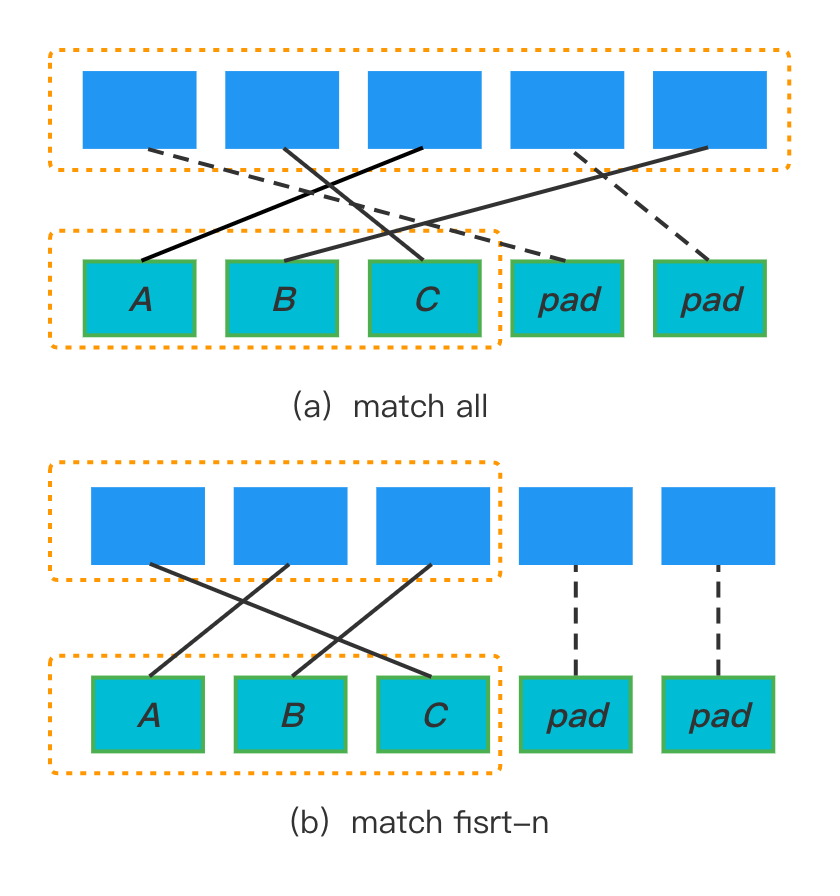}
    \caption{Two assignment schemes.}
    \label{fig: two bipartite matching scheme}
\end{figure}

In practice, we design two assignment schemes for bipartite matching. The first scheme is to get the optimal matching between non-$\varnothing$ ground truth labels and all generated distributions, then assign $\varnothing$ labels to the rest generated distributions one-to-one. The second scheme is to get the optimal matching between non-$\varnothing$ ground truth labels and first-n generated distributions, then assign $\varnothing$ labels to the rest generated distributions one-to-one. Figure \ref{fig: two bipartite matching scheme} shows the two assignment schemes.

Finally, we get the bipartite matching loss based on the reordered sequence $\boldsymbol{\hat{y}}=\{y_{\hat{\rho}(i)}\}_{i=1,\ldots,N}$ to train model, which is defined as follows:
\begin{multline} \label{eq:bipartite}
    \mathcal{L}_{b}(\boldsymbol{\hat{y}},\boldsymbol{p})=-\sum_{i=1}^{N}[\mathbbm{1}_{\{y_{\hat{\rho}(i)}\neq\varnothing\}}\log\boldsymbol{p}_{i}(y_{\hat{\rho}(i)}) \\ 
    +\mathbbm{1}_{\{y_{\hat{\rho}(i)}=\varnothing\}}\lambda_{\varnothing}\log\boldsymbol{p}_{i}(y_{\hat{\rho}(i)})]
\end{multline}

where $\lambda_{\varnothing}$ is a scale factor less than $1$ that forces the model concentrate more on non-$\varnothing$ labels.
\subsubsection{Semantic Optimal Transport Distance}
\begin{figure}[t]
    \centering
    \includegraphics[width=7.5cm]{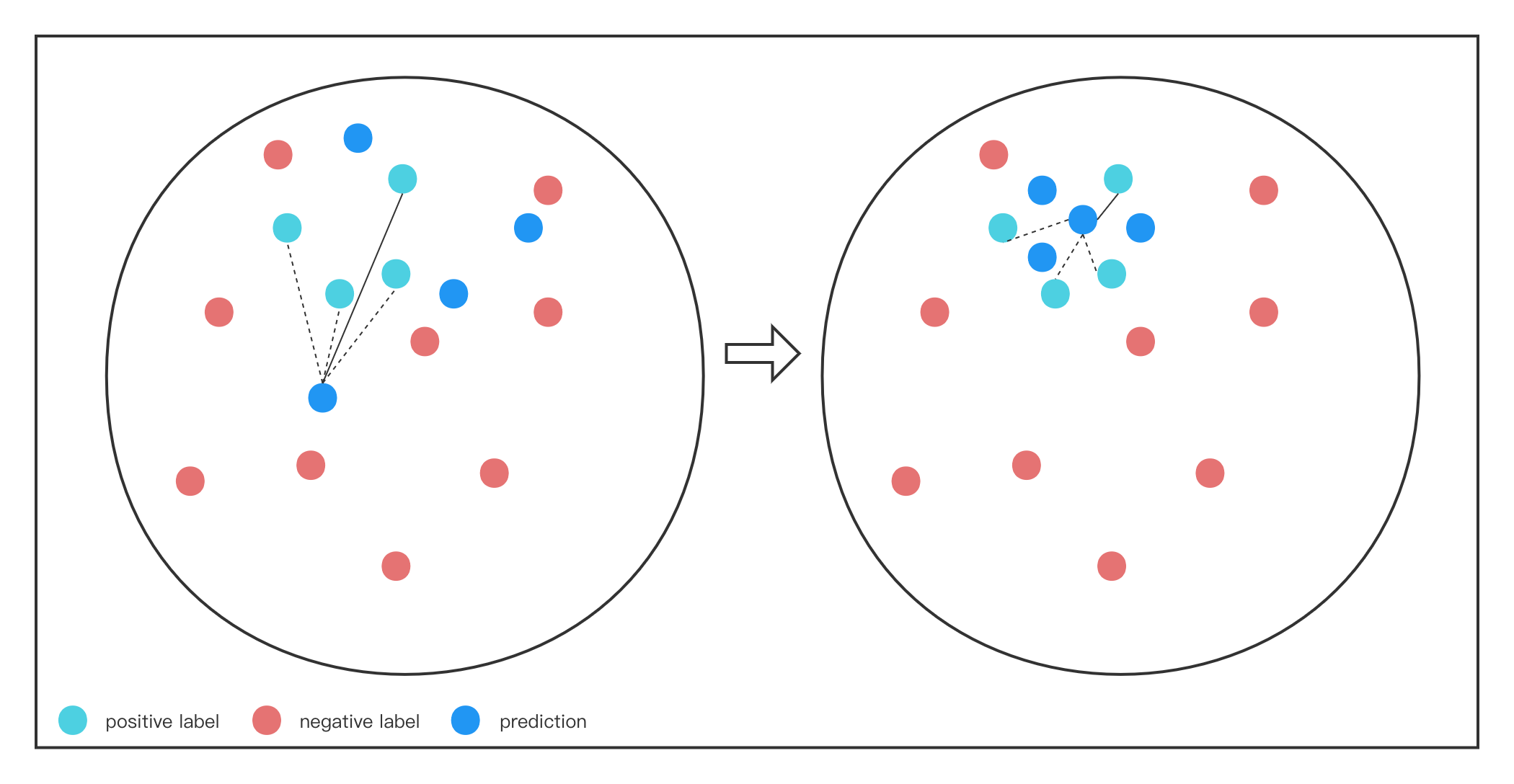}
    \caption{Semantic optimal transport distance in embedding space.}
    \label{fig: OT in embedding space}
\end{figure}

In XMTC, semantic similar labels commonly appear together in each sample. The bipartite matching loss described in Eq.(\ref{eq:bipartite}) only utilizes one value in each predictions. We completely utilize the predictions in training, and take the optimal transport (OT) distance in embedding space as a regularization term to make all predictions close to positive labels, as the shown in Figure \ref{fig: OT in embedding space}.
\paragraph{OT distance on discrete domain}
The OT distance is also known as Wasserstein distance on a discrete domain $\mathbbm{X}$ (the sequence space), which is defined as follows:
\begin{multline}
    \mathcal{L}_{ot}(\boldsymbol{p},\boldsymbol{y})=\min_{\boldsymbol{\Gamma}\in\Sigma(\boldsymbol{\mu},\boldsymbol{\nu})}\langle\boldsymbol{C},\boldsymbol{\Gamma}\rangle\\
    =\min_{\boldsymbol{\Gamma}\in\Sigma(\boldsymbol{\mu},\boldsymbol{\nu})}\sum_{i=1}^n\sum_{j=1}^m\Gamma_{ij}\cdot c(\boldsymbol{p}_i, y_j),
\end{multline}
where $\boldsymbol{\mu}\in \mathbbm{R}_{+}^n$, $\boldsymbol{\nu}\in \mathbbm{{R}_{+}^m}$ are two discrete distributions on $\mathbbm{X}$, $\|\boldsymbol{\mu}\|=\|\boldsymbol{\nu}\|=1$. In our case, given realizations $\{\boldsymbol{p}_i\}_{i=1}^n$ and $\{y_j\}_{j=1}^m$ of $\boldsymbol{\mu}$ and $\boldsymbol{\nu}$, we can approximate them by empirical distributions $\boldsymbol{\widehat{\mu}}=\frac{1}{n}\sum_{\boldsymbol{p}_i}\delta_{\boldsymbol{p}_i}$ and $\boldsymbol{\widehat{\nu}}=\frac{1}{m}\sum_{y_j}\delta_{y_j}$. The supports of $\boldsymbol{\widehat{\mu}}$ and $\boldsymbol{\widehat{\nu}}$ are finite, so finally we have $\boldsymbol{\mu}=\frac{1}{n}\boldsymbol{1}_{\{\boldsymbol{p}_i\}}$ and $\boldsymbol{\nu}=\frac{1}{m}\boldsymbol{1}_{\{y_j\}}$. $\Sigma(\boldsymbol{\mu},\boldsymbol{\nu})$ is the set of joint distributions whose marginals are $\boldsymbol{\mu}$ and $\boldsymbol{\nu}$, which is described as $\Sigma(\boldsymbol{\mu},\boldsymbol{\nu})=\{\boldsymbol{\Gamma}\in \mathbbm{R}_{+}^{n\times m}|\boldsymbol{\Gamma} \boldsymbol{1}_m=\boldsymbol{\mu},\boldsymbol{\Gamma^\top} \boldsymbol{1}_n=\boldsymbol{\nu}\}$, where $\boldsymbol{1}_n$ represents n-dimensional vector of ones. Matrix $\boldsymbol{C}=[c(\boldsymbol{p}_i,y_j)]\in \mathbbm{R}_{+}^{n\times m}$ is the cost matrix, whose element $c(\boldsymbol{p}_i,y_j)$ denotes the distance between the $i$-th support point $\boldsymbol{p}_i\in \mathbbm{X}$ of $\boldsymbol{\mu}$ and the $j$-th support point $y_j\in \mathbbm{X}$ of $\boldsymbol{\nu}$. Notation $\langle\cdot,\cdot\rangle$ represents the Frobenius dot-product. With the optimal solution $\boldsymbol{\Gamma}^*$, $\mathcal{L}_{ot}(\boldsymbol{p},\boldsymbol{y})$ is the minimum cost that transport from distribution $\boldsymbol{\mu}$ to $\boldsymbol{\nu}$.

We use a robust and efficient iterative algorithm to compute the OT distance, which is the recently introduced Inexact Proximal point method for exact Optimal Transport (IPOT) \citep{xie2020fast}. 


\paragraph{Semantic Cost Function in OT}
In XMTC learning, it's intractable to directly compute the cost distance between predictions and ground-truth one-hot vector because of the huge label space. A more natural choice for computing cost distance is to use the label embeddings. Based on the rich semantics of embedding space, we can compute the semantic OT distance between model predictions and ground-truth. The details of computing the \textit{cosine distance} are described as follows:
\begin{equation}
    c(\boldsymbol{p}_i, y_j) = cosine\_distance(\boldsymbol{E}^{\top}\boldsymbol{p}_i, \boldsymbol{E}^{\top}\mathbbm{1}_{y_j})
\end{equation}
\begin{equation}
    cosine\_distance(\boldsymbol{x}, \boldsymbol{y})= 1 - \frac{\boldsymbol{x}^{\top}\boldsymbol{y}}{\|\boldsymbol{x}\|_{2}\|\boldsymbol{y}\|_2}
\end{equation}
where $\boldsymbol{E}\in \mathbbm{R}^{V\times d}$ is the label embedding matrix, $V$ is the vocabulary size and $d$ is the dimension for the embedding vector. $\mathbbm{1}_{y_j}$ is the one-hot vector whose value is 0 in all positions except at position $y_j$ is 1.

\begin{figure}[t]
    \centering
    \includegraphics[width=7.5cm]{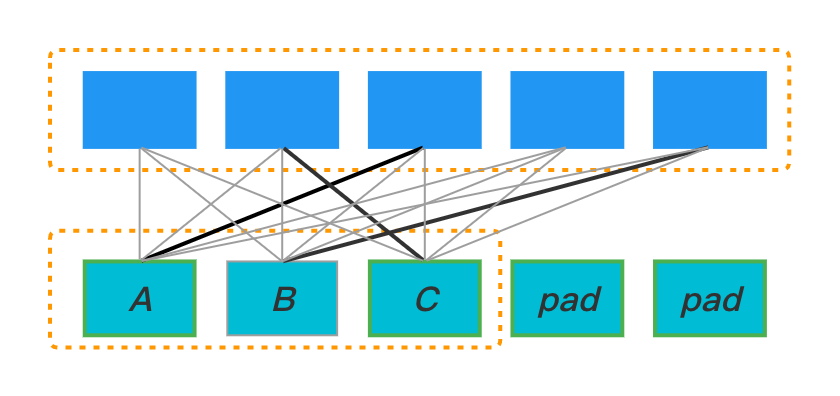}
    \caption{Semantic optimal transport distance.}
    \label{fig: Semantic OT distance}
\end{figure}

In practice, we compute the semantic optimal transport distance $\mathcal{L}_{ot}(\boldsymbol{p},\boldsymbol{y})$ between non-$\varnothing$ ground-truth labels and all predictions, which is shown in \ref{fig: Semantic OT distance}.

\subsubsection{Complete Seq2Set Training with OT regularization}
The bipartite matching gives the optimal matching between ground truth and predictions, which enables the Seq2Set model to train with Maximum likelihood estimation (MLE). In addition, the OT distance provides a measurement in semantic label embedding space. We propose to combine the OT distance and the bipartite matching loss, which gives us the final training objective:
\begin{equation}\label{eq:loss}
    \mathcal{L} = \mathcal{L}_{b}(\boldsymbol{\hat{y}},\boldsymbol{p}) + \lambda \mathcal{L}_{ot}(\boldsymbol{p},\boldsymbol{y})
\end{equation}
where $\lambda$ is a hyper-parameter to be tuned.

\begingroup
\setlength{\tabcolsep}{3.8pt}
\begin{table*}[ht]
    \centering
    \begin{tabular}{l| r r r r r r r r r}
        \hline
        Dataset & $N_{train}$ & $N_{test}$ & $D$ & $L$ & $\overline{L}$ & $L_{max}$ & $\hat{L}$ & $\overline{W}_{train}$ & $\overline{W}_{test}$ \\
        \hline
         Eurlex-4K & 15,449 & 3,865 & 186,104 & 3,956 & 5.30 & 24 & 20.79 & 1248.58 & 1230.40 \\
        Wiki10-31K & 14,146 & 6,616 & 101,938 & 30,938 & 18.64 & 30 & 8.52 & 2484.30 & 2425.45 \\
        AmazonCat-13K & 1,186,239 & 306,782 & 203,882 & 13,330 & 5.04 & 57 & 448.57 & 246.61 & 245.98 \\
        Amazon-670K & 490,449 & 153,025 & 135,909 & 670,091 & 5.45 & 7 & 3.99 & 247.33 & 241.22 \\
        \hline
    \end{tabular}
    \caption{Data Statistics. $N_{train}$ and $N_{test}$  refer to the number of training samples and test samples respectively. $D$ refers to the total number of features. $L$ is the total number of labels, $\overline{L}$ is the average number of labels per training sample, $L_{max}$ is the max length of labels for both training and test samples, $\hat{L}$ is the average number of samples per training label. $\overline{W}_{train}$ and $\overline{W}_{test}$ refer to the average number of words per training and test sample respectively.}
    \label{tab:datasets}
\end{table*}
\endgroup

\section{Experiments}
\subsection{Datasets}
We use four benchmark datasets, including three large-scale datasets Eurlex-4K \citep{loza_mencia_efficient_2008} and AmazonCat-13K \citep{mcauley_hidden_2013} and Wiki10-31K \citep{zubiaga_enhancing_2012}, one extreme-scale dataset Amazon-670K \citep{mcauley_hidden_2013}. We follow the training and test split of AttentionXML \citep{you_attentionxml_2019} and we set aside 10\% of the training instances as the validation set for hyperparameter tuning. The dataset statistics are summarized in Table \ref{tab:datasets}.
\subsection{Evaluation Metrics}
Following the previous work (\citealp{yang_sgm_2018}, \citealp{nam_maximizing_2017}, \citealp{yang_deep_2019}), we evaluate all methods in terms of micro-F1 score, micro-precision and micro-recall are also reported for analysis.
\subsection{Baselines}
We compare our proposed methods with the following competitive baselines:

\textbf{Seq2Seq} \citep{bahdanau_neural_2016} is a classic Seq2Seq model with attention mechanism, which is also a strong baseline for XMTC tasks.

\textbf{MLC2Seq} \citep{nam_maximizing_2017} is based on Seq2Seq \citep{bahdanau_neural_2016}, which enhances Seq2Seq model for MLC tasks by changing label permutations according to the frequency of labels. We take the descending order from frequent to rare in our experiments.

\textbf{SGM} \citep{yang_sgm_2018} is based on previews Seq2Seq \citep{bahdanau_neural_2016} and MLC2Seq \citep{nam_maximizing_2017}, which views the MLC task as a sequence generation problem to take the correlations between labels into account.

In order to extend the above three models to large-scale dataset Wiki10-31K and extreme-scale dataset Amazon-670K, we use the bottleneck layer to replace the fully connected layer. The bottleneck sizes are same as our proposed model and they are shown in Table \ref{tab:hyperparameters}. 

\subsection{Experiment Settings}
For each dataset, the vocabulary size of input text is limited to 500,000 words according to the word frequency in the dataset. 

The number of GRU layers of the encoder is 2, and for the decoder is 1. The hidden sizes of the encoder and the decoder both are 512. We set the max length of generated predictions $N$ to the max length of labels in each dataset are shown in Table \ref{tab:datasets} as $L_{max}$.

For LightConv, we set the kernel sizes to 3,7,15,30 for each layer respectively. To reduce the memory consumption of GPU, we set the stride of convolution to 3 in the last layer for all datasets except that Wiki10-31K is 4.

We set the hyperparameter $\lambda_{\varnothing}$ of bipartite matching loss to 0.2 in Eq.(\ref{eq:bipartite}) following \citet{ye_one2set_2021}, and $\lambda$ of final loss to 8 in Eq.(\ref{eq:loss}).

For word embeddings in three baseline models and our proposed models, we use pre-trained 300-dimensional Glove vectors \citep{pennington_glove_2014} to all datasets for input text, and use the mean value of embeddings of words in labels for all datasets except for Amazon-670K, because the labels in Amazon-670K are corresponding item numbers and we randomly initialize 100-dimensional embeddings for Amazon-670K.

All models are trained by the Adam optimizer \citep{kingma2014adam} with a cosine annealing schedule. Besides, we use dropout \citep{srivastava2014dropout} to avoid overfitting, the dropout rate is 0.2, and clip the gradients \citep{pascanu2013difficulty} to the maximum norm of 8.

All models are trained on one Nvidia TITAN V and one Nvidia GeForce RTX 3090, for the small dataset Eurlex-4K, we use TITAN V, and for others, we use RTX 3090.

Other hyperparameters are given in Table \ref{tab:hyperparameters}.

\begingroup
\setlength{\tabcolsep}{4pt}
\begin{table}[h]
    \centering
    \begin{tabular}{c|c c c c c}
    \hline
    Datasets & $E$ & $B$ & $lr$ & $b$ & $L_{t}$ \\
    \hline
    Eurlex-4K & 20 & 32 & 0.001 & - & 1000  \\
    Wiki-31K & 15 & 32 & 0.001 & 300 & 2000 \\
    AmazonCat-13K & 5 & 64 & 0.0005 & -  & 500 \\
    Amazon-670K & 20 & 64 & 0.0005 & 512 & 500 \\
    \hline
    \end{tabular}
    \caption{Hyperparameters of all datasets. $E$ is the number of epochs, $B$ is the batch size, $lr$ is init learning rate, $b$ is the bottleneck size (Note that the bottleneck layer is not used on Eurlex-4K and AmanzonCat-13K), $L_{t}$ is the maximum length of input tokens.}
    \label{tab:hyperparameters}
\end{table}
\endgroup

\begin{table*}[ht]
    \centering
    \begin{tabular}{c c c c}
        \hline
        Methods & P & R & F1  \\
        \hline
        \hline
        \multicolumn{4}{c}{Eurlex-4K} \\
        \hline
        Seq2Seq & 60.83 & 54.76 & 57.64 \\
        MLC2Seq & 59.80 & 56.70 & 58.21 \\
        SGM & 59.99 & 56.56 & 58.23 \\
        \hline
        Seq2Seq + LC & 61.36 & 55.03 & 58.02 \\
        \hline
        $\mathrm{BM}_{all}$ & 61.72 & 57.63 & 59.61 \\
        $\mathrm{BM}_{all}$ + LC & 63.38 & 58.10 & 60.63 \\
        $\mathrm{BM}_{all}$ + OT & \textbf{65.95} & 58.54 & 62.03 \\
        $\mathrm{BM}_{all}$ + OT + LC & 64.82 & \textbf{60.26} & \textbf{62.46} \\
        \hline
    \end{tabular}
    \begin{tabular}{c c c c}
        \hline
        Methods & P & R & F1  \\
        \hline
        \hline
        \multicolumn{4}{c}{Wiki10-31K} \\
        \hline
        Seq2Seq & 29.87 & 26.96 & 28.34 \\
        MLC2Seq & 28.69 & 28.03 & 28.36 \\
        SGM & 28.89 & 27.93 & 28.40 \\
        \hline
        Seq2Seq + LC & 30.06 & 27.14 & 28.53 \\ 
        \hline
        $\mathrm{BM}_{first-n}$ & 33.23 & 29.43 & 31.22 \\
        $\mathrm{BM}_{first-n}$ + LC & 33.58 & 29.94 & 31.66 \\
        $\mathrm{BM}_{first-n}$ + OT & \textbf{37.26} & 29.08 & 32.67 \\
        $\mathrm{BM}_{first-n}$ + OT + LC & 36.70 & \textbf{30.05} & \textbf{33.04} \\
        \hline
    \end{tabular}
    \begin{tabular}{c c c c}
        \hline
        \multicolumn{4}{c}{Amazon-670K} \\
        \hline
        Seq2Seq & 32.16 & 32.04 & 32.10 \\
        MLC2Seq & 30.27 & \textbf{32.32} & 31.26 \\
        SGM & 30.12 & 31.42 & 30.76 \\
        \hline
        Seq2Seq + LC & 31.15 & 31.32 & 31.24 \\
        \hline
        $\mathrm{BM}_{all}$ & 55.23 & 27.33 & 36.56 \\
        $\mathrm{BM}_{all}$ + LC & 52.70 & 26.30 & 35.09 \\
        $\mathrm{BM}_{all}$ + OT & 58.36 & 26.46 & 36.42 \\
        $\mathrm{BM}_{all}$ + OT + LC & \textbf{61.16} & 26.39 & \textbf{36.87} \\
        \hline
    \end{tabular}
    \begin{tabular}{c c c c}
        \hline
        \multicolumn{4}{c}{AmazonCat-13K} \\
        \hline
        Seq2Seq & 74.26 & 68.24 & 71.12 \\
        MLC2Seq & 70.93 & 69.66 & 70.29 \\
        SGM & 71.27 & 69.36 & 70.30 \\
        \hline
        Seq2Seq + LC & \textbf{74.60} & 68.49 & 71.41 \\
        \hline
        $\mathrm{BM}_{first-n}$ & 70.26 & 72.51 & 71.37 \\
        $\mathrm{BM}_{first-n}$ + LC & 70.46 & \textbf{72.95} & 71.68 \\
        $\mathrm{BM}_{first-n}$ + OT & 71.80 & 71.57 & 71.69 \\
        $\mathrm{BM}_{first-n}$ + OT + LC & 71.98 & 71.71 & \textbf{71.84} \\
        \hline
    \end{tabular}
    \caption{Comparison between our models and three baselines on Eurlex-4K, Wiki10-31K, Amazon-670K, and AmazonCat-13K. $\mathrm{BM}_{all}$ denotes the Seq2Seq model with the first scheme of bipartite matching loss, $\mathrm{BM}_{first-n}$ denotes the Seq2Seq model with the second one. OT denotes the semantic optimal transport distance loss. LC denotes the light weight convolution layer. We take the average score of 4 times experiments for all models on Eurlex-4K and Wiki10-31K, and 2 times experiments for the other two datasets.}
    \label{tab:results}
\end{table*}

\subsection{Main Results}
Table \ref{tab:results} compares the proposed methods with three baselines on four benchmark datasets. We focus on micro-F1 score following previous works based on Seq2Seq. The best score of each metric is in boldface. Our models outperform all baselines in micro-F1 score. We find that the two assignment scheme of bipartite matching each excel on different datasets. $\mathrm{BM}_{all}$ denotes the Seq2Seq model with the first scheme of bipartite matching, $\mathrm{BM}_{first-n}$ denotes the Seq2Seq model with the second one. $\mathrm{BM}_{all}$ has the better performance on Eurlex-4K and Amazon-670K, rather $\mathrm{BM}_{first-n}$ is better on Wiki10-31K and AmazonCat-13K. Our complete method achieves a large improvement of 16.34\% micro-F1 score over the second best baseline model which is SGM on Wiki10-31K, and 14.86\%, 7.26\%, 1.01\% on other datasets respectively, the relative impovements are shown in Table \ref{tab:improvements}.

\begin{table*}[h]
    \centering
    \begin{tabular}{c c c c c}
    \hline
    Methods & Eurlex-4K & Wiki10-31K & Amazon-670K & AmazonCat-13K \\
    \hline
    $\mathrm{BM}$ & 2.37\% & 9.93\% & 13.89\% & 0.35\% \\
    $\mathrm{BM}$ + OT & 6.53\% & 15.04\% & 13.46\% & 0.8\% \\
    $\mathrm{BM}$ + OT + LC & 7.26\% & 16.34\% & 14.86\% & 1.01\% \\
    \hline
    \end{tabular}
    \caption{Relative improvements of micro-F1 score over the second best baseline model on four datasets in Table \ref{tab:results}. $\mathrm{BM}$ denotes $\mathrm{BM}_{all}$ or $\mathrm{BM}_{first-n}$.}
    \label{tab:improvements}
\end{table*}

\subsection{$\boldsymbol{\mathrm{BM}_{first-n}}$ vs $\boldsymbol{\mathrm{BM}_{all}}$}
\begin{figure}
    \centering
    \includegraphics[width=8.5cm]{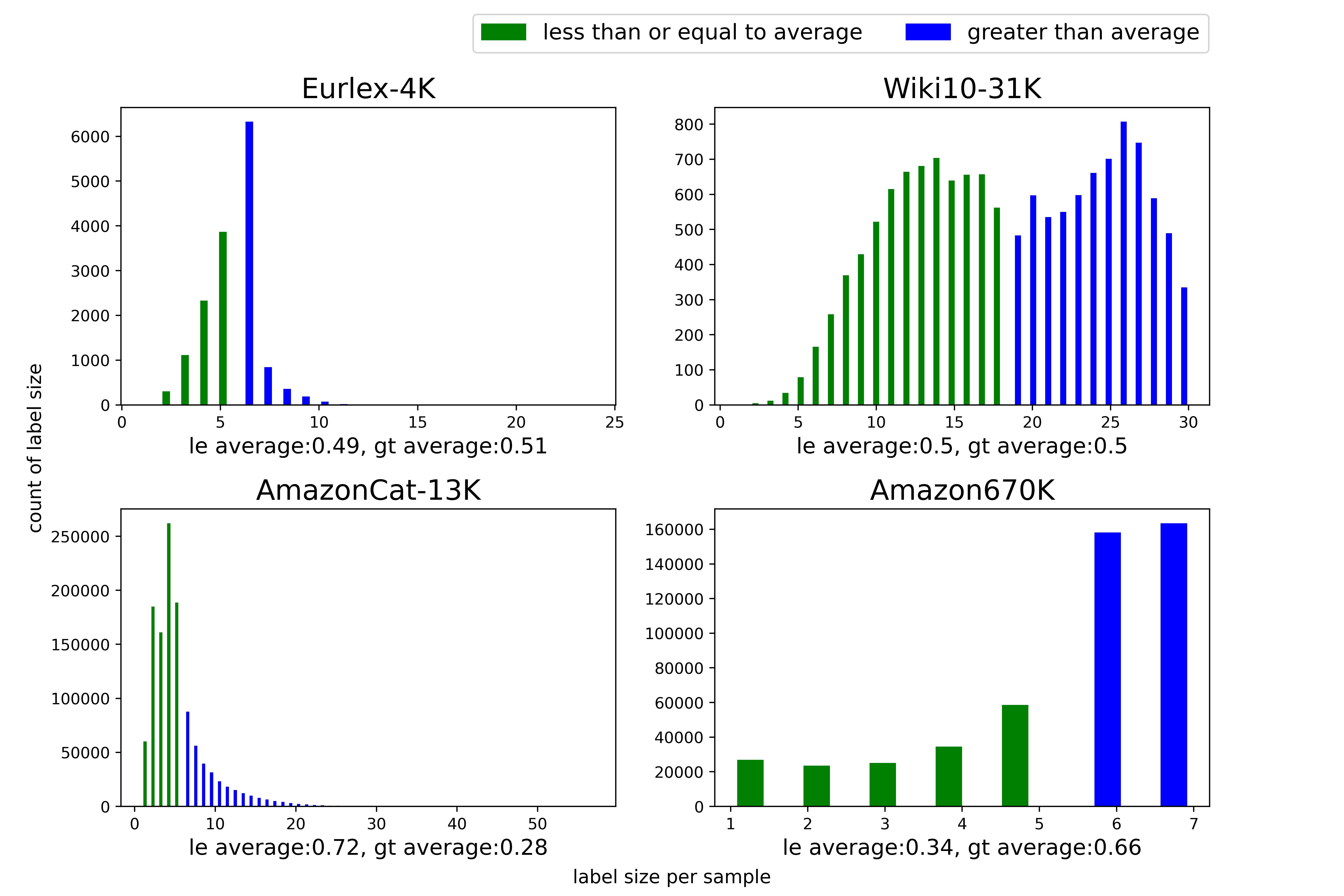}
    \caption{Distributions of label size on all datasets. The \textit{le average} denotes the proportion of samples whose label size less than or equal to average number of label size, the \textit{gt average} denotes the opposite proportion.}
    \label{fig:label_size}
\end{figure}

We find that the different performance of $\mathrm{BM}_{first-n}$ and $\mathrm{BM}_{all}$ are related to the distribution of label size on the dataset. Table \ref{tab:all vs first-n} compares $\mathrm{BM}_{first-n}$ with $\mathrm{BM}_{all}$ in micro-F1 score on all datasets. The performance difference on Eurlex-4K is very small since the two proportions which are shown in Figure \ref{fig:label_size} are nearly equal. Wiki10-31K is the same case. However, the two proportions have a big difference on AmanzonCat-13K and Amazon-670K, which leads to the large performance difference between $\mathrm{BM}_{first-n}$ and $\mathrm{BM}_{all}$. The proportion of samples whose label size is smaller than the average is less than 50\%, then the performance of $\mathrm{BM}_{first-n}$ is better than $\mathrm{BM}_{all}$. The proportion of samples whose label size is greater than average number is greater than 50\%, then the performance of $\mathrm{BM}_{all}$ is better than $\mathrm{BM}_{first-n}$.
\begin{table}
    \centering
    \begin{tabular}{c c c}
    \hline
    Datasets & $\mathrm{BM}_{all}$ & $\mathrm{BM}_{first-n}$ \\
    \hline
    Eurlex-4K & 59.61 & 59.12($\downarrow 0.82\%$) \\
    Wiki10-31K & 31.38 & 31.22($\downarrow 0.51\%$) \\
    AmazonCat-13K & 67.09 & 71.37($\uparrow 6.38\%$) \\
    Amazon-670K & 36.87 & 34.45($\downarrow 6.56\%$) \\
    \hline
    \end{tabular}
    \caption{The micro-F1 score of $\mathrm{BM}_{all}$ and $\mathrm{BM}_{first-n}$ and relative difference on four datasets.}
    \label{tab:all vs first-n}
\end{table}

\subsection{Effect of Lightweight Convolution and Semantic Optimal Transport Distance}
To examine the effect of LightConv and OT, we add LightConv to the Seq2seq model and $\mathrm{BM}$ model, and add OT to $\mathrm{BM}$ model. The details of results are shown in Table \ref{tab:results}.
The semantic optimal transport distance and LigntConv achieve improvements against origin models on all datasets except for Amazon-670K, but it's not beyond our imagination, because Amazon-670K has an extremely large label set which is item numbers. The item numbers don't have semantic information leading to OT having no effect. At the same time Amazon-670K has the lowest average number of samples per label, it's very hard to learn a proper context vector by the LightConv. However, the combination of OT and LightConv still has a small improvement over $\mathrm{BM}_{all}$ methods on Amazon-670K.

\subsection{Comprehensive Comparison}
To realize the performance of our method on tail labels, we use the macro-averaged F1 (maF1) score which treats all labels equally regardless of their support values. To more comprehensively compare our model with baselines, we also use the weighted-averaged F1 (weF1) which considers each label's support, and the example-based F1 (ebF1) which calculates F1 score for each instance and finds their average. We show the results in Table \ref{tab:3f1}.

\begin{table*}[ht]
    \centering
    \begin{tabular}{c c c c}
        \hline
        Methods & maF1 & weF1 & ebF1  \\
        \hline
        \hline
        \multicolumn{4}{c}{Eurlex-4K} \\
        \hline
        Seq2Seq & 23.26 & 55.36 & 57.10 \\
        MLC2Seq & 24.16 & 56.24 & 58.00 \\
        SGM & 24.32 & 56.28 & 57.96 \\
        $\mathrm{OTSeq2Set}_{all}$ & \textbf{27.34} & \textbf{60.45} & \textbf{61.04} \\
        \hline
    \end{tabular}
    \begin{tabular}{c c c c}
        \hline
        Methods & maF1 & weF1 & ebF1  \\
        \hline
        \hline
        \multicolumn{4}{c}{Wiki10-31K} \\
        \hline
        Seq2Seq & 3.51 & 24.55 & 27.97 \\
        MLC2Seq & \textbf{4.85} & 26.15 & 28.14 \\
        SGM & 4.29 & 26.00 & 28.19 \\
        $\mathrm{OTSeq2Set}_{first-n}$ & 3.50 & \textbf{26.55} & \textbf{32.65} \\
        \hline
    \end{tabular}
    \begin{tabular}{c c c c}
        \hline
        \multicolumn{4}{c}{Amazon-670K} \\
        \hline
        Seq2Seq & \textbf{11.57} & 30.29 & \textbf{29.62} \\
        MLC2Seq & 11.55 & \textbf{30.36} & \textbf{29.62} \\
        SGM & 11.14 & 29.53 & 28.85 \\
        $\mathrm{OTSeq2Set}_{all}$ & 9.91 & 28.83 & 25.65 \\
        \hline
    \end{tabular}
    \begin{tabular}{c c c c}
        \hline
        \multicolumn{4}{c}{AmazonCat-13K} \\
        \hline
        Seq2Seq & 41.28 & 69.37 & 74.03 \\
        MLC2Seq & 39.79 & 68.78 & 73.57 \\
        SGM & 40.44 & 68.96 & 73.68 \\
        $\mathrm{OTSeq2Set}_{first-n}$ & \textbf{43.42} & \textbf{70.95} & \textbf{75.18} \\
        \hline
    \end{tabular}
    \caption{Performance comparison of all models on the macro-averaged F1, the weighted-
averaged F1 and the example-based F1.}
    \label{tab:3f1}
\end{table*}

\subsection{Performance over different $\lambda$}
\begin{figure}[h]
    \centering
    \includegraphics[width=7.5cm]{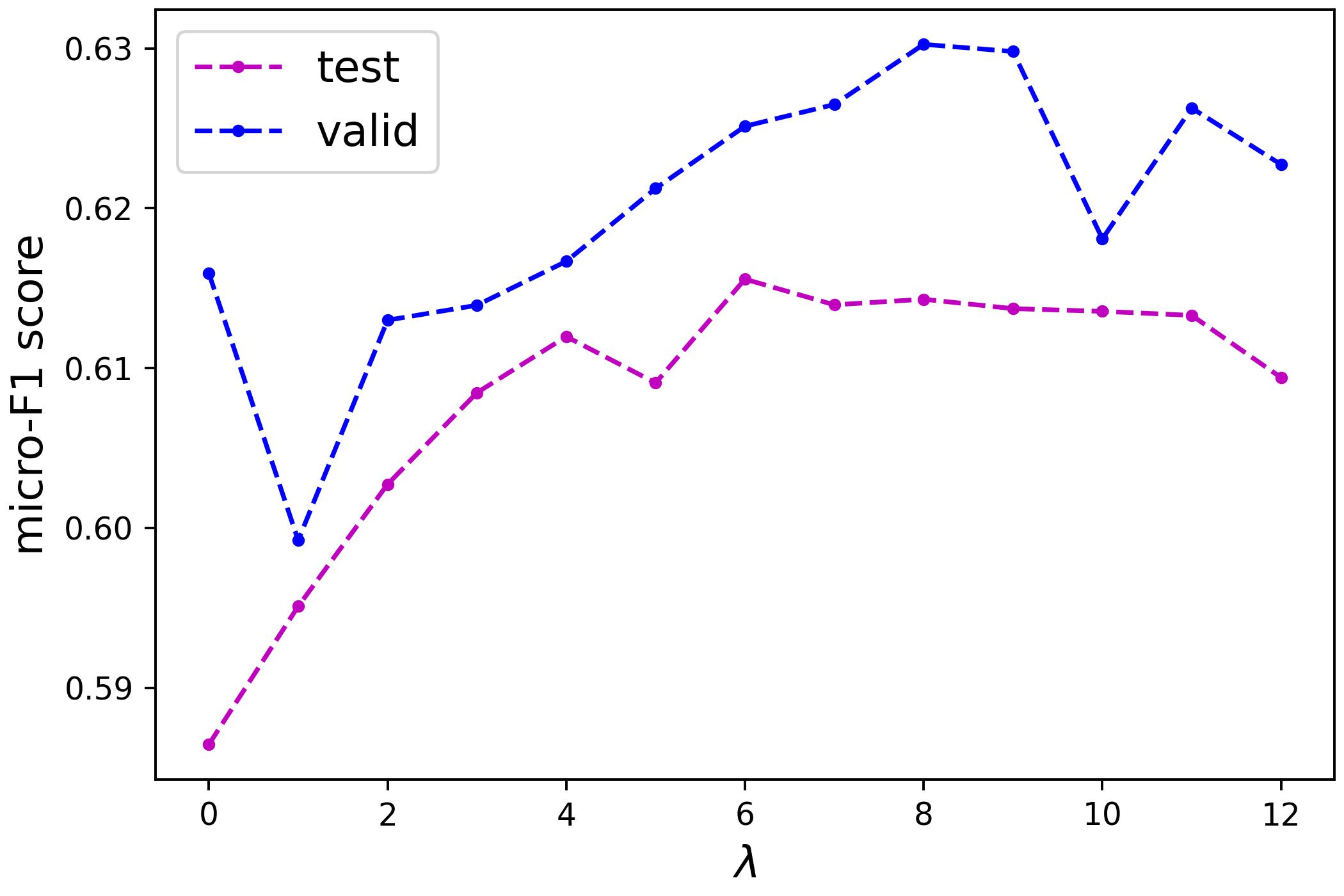}
    \caption{Performance of the OTSeq2Set using different $\lambda$ in loss function.}
    \label{fig:lambda}
\end{figure}

We conduct experiments on Eurlex4K dataset to evaluate performance with different hyperparameter $\lambda$ in complete loss function $\mathcal{L}$. The results are shown in Figure \ref{fig:lambda}. The figure shows that when $\lambda =8$, the performance reaches its best and is stable.

\subsection{Computation Time and Model Size}
Table \ref{tab:time and size} shows the overall training time of all models compared in our experiments. The MLC2Seq has the same model size and training time as Seq2Seq.
The model size of OTSeq2Set is almost the same as that of Seq2Seq, but the overall training time is much longer than other models, because the bipartite matching loss and the optimal transport distance need to compute for each sample individually.

\begingroup
\setlength{\tabcolsep}{2pt}
\begin{table}[]
    \centering
    \begin{tabular}{l c r r r}
    \hline
    Datasets & & Seq2Seq & SGM & OTSeq2Set \\
    \hline
    \multirow{2}{4em}{Eurlex-4K} & $T_{train}$ & 0.85 & 0.85 & 1.52 \\
    & $M$ & 34.8 & 35.0 & 40.5 \\
    \multirow{2}{4em}{Wiki10-31K} & $T_{train}$ & 1.55 & 1.56 & 1.78 \\
    & $M$ & 45.5 & 45.7 & 47.5 \\
    \multirow{2}{5em}{AmazonCat-13K} & $T_{train}$ & 3.70 & 3.75 & 17.88 \\
    & $M$ & 54.8 & 55.0 & 70.1 \\
    \multirow{2}{4em}{Amazon-670K} & $T_{train}$ & 12.9 & 16.0 & 18.5 \\
    & $M$ & 437.7 & 437.7 & 439.8 \\
    \hline
    \end{tabular}
    \caption{Training Time and Model Size. $T_{train}$ is the overall training hours. $M$ is the total size of model parameters in millions.}
    \label{tab:time and size}
\end{table}
\endgroup

\section{Related Work}
The most popular deep learning models in XMTC are fully connected layer based models. XML-CNN \citep{liu_deep_2017} uses a convolutional neural network (CNN) and a dynamic max pooling scheme to learn the text representation, and adds an hidden bottleneck layer to reduce model size as well as boost model performance. AttentionXML \citep{you_attentionxml_2019} utilizes a probabilistic label tree (PLT) to handle millions of labels. In AttentionXML, a BiLSTM is used to capture long-distane dependency among words and a multi-label attention is used to capture the most relevant parts of texts to each label. CorNet \citep{xun_correlation_2020} proposes a correlation network (CorNet) capable of ulilizing label correlations. X-Transformers \citep{chang_Taming_2020} proposes a scalable approach to fine-tuning deep transformer models for XMTC tasks, which is also the first method using deep transformer models in XMTC. LightXML \citep{jiang_lightxml_2021} combines the transformer model with generative cooperative networks to fine-tune transformer model.

Another type of deep learning-based method is Seq2Seq learning based methods. MLC2Seq \citep{nam_maximizing_2017} is based on the classic Seq2Seq \citep{bahdanau_neural_2016} architecture which uses a bidirectional RNN to encode raw text and an RNN with an attention mechanism to generate predictions sequentially. MLC2Seq enhances the model performance by determining label permutations before training. SGM \citep{yang_sgm_2018} proposes a novel decoder structure to capture the correlations between labels. \citet{yang_deep_2019} proposes a two-stage training method, which is trained with MLE then is trained with a self-critical policy gradient training algorithm.

About bipartite matching, \citet{tan_sequence--set_2021} considers named entity recognition as a Seq2Set task, the model generates an entity set by a non-autoregressive decoder and is trained by a loss function based on bipartite matching. ONE2SET \citep{ye_one2set_2021} proposes a K-step target assignment mechanism via bipartite matching on the task of keyphrases generation, which also uses a non-autoregressive decoder.

\citet{xie2020fast} proposes a fast proximal point method to compute the optimal transport distance, which is named as IPOT. \citet{chen_improving_2019} adds the OT distance as a regularization term to the MLE training loss. The OT distance aims to find an optimal matching of similar words/phrases between two sequences which promotes their semantic similarity. \citet{li_improving_2020} introduce a method that combines the student-forcing scheme with the OT distance in text generation tasks, this method can alleviate the exposure bias in Seq2Seq learning. The above two methods both use the IPOT to compute the OT distance.

\section{Conclusion}
In this paper, we propose an autoregressive Seq2Set model for XMTC, OTSeq2Set, which combines the bipartite matching and the optimal transport distance to compute overall training loss and uses the student-forcing scheme in the training state. OTSeq2Set not only eliminates the influence of order in labels, but also avoids the exposure bias. Besides, we design two schemes for the bipartite matching which are suitable for datasets with different label distributions. The semantic optimal transport distance can enhance the performance by the semantic similarity of labels. To take full advantage of the raw text information, we add a lightweight convolution module which achieves a stable improvement and requires only a few parameters. Experiments show that our method gains significant performance improvements against strong baselines on XMTC. 

\section*{Limitations}
For better effect, we compute the bipartite matching and the optimal transport distance between non-$\varnothing$ targets and predictions. We can't use batch computation to improve efficiency so that the training time of OTSeq2Set is longer than other baseline models.

\section*{Acknowledgement}
This study is supported by China Knowledge Centre for Engineering Sciences and Technology (CKCEST).
\bibliography{custom}
\bibliographystyle{acl_natbib}

\appendix

\end{document}